\documentclass[11pt]{article}
\usepackage{geometry}
\usepackage{coling2020}
\usepackage{times}
\usepackage{url}
\usepackage{latexsym}
\usepackage{microtype}
\usepackage{color}
\usepackage{multirow}
\usepackage{colortbl}
\usepackage{subcaption}
\usepackage{arydshln}

\colingfinalcopy 

\title{UoB at SemEval-2020 Task 12: Boosting BERT with Corpus Level Information.}

\author{Wah Meng Lim \\
  University of Birmingham \\
  United Kingdom \\
    {\tt \small  wahmengsch@gmail.com} \\\And
Harish Tayyar Madabushi \\
  University of Birmingham \\
  United Kingdom \\
  {\tt \small H.TayyarMadabushi.1@bham.ac.uk} \\ }

\date{}

\begin{document}
\maketitle
\begin{abstract}


Pre-trained language model word representation, such as BERT, have been extremely successful in several Natural Language Processing tasks significantly improving on the state-of-the-art. This can largely be attributed to their ability to better capture semantic information contained within a sentence. Several tasks, however, can benefit from information available at a corpus level, such as Term Frequency-Inverse Document Frequency (TF-IDF). In this work we test the effectiveness of integrating this information with BERT on the task of identifying abuse on social media and show that integrating this information with BERT does indeed significantly improve performance. We participate in Sub-Task A (abuse detection) wherein we achieve a score within two points of the top performing team and in Sub-Task B (target detection) wherein we are ranked 4 of the 44 participating teams.
\end{abstract}

\section{Introduction}

\blfootnote{
    %
    %
    %
    %
    \hspace{-0.65cm}  
    Accepted for publication in the Proceedings of the 14\textsuperscript{th}
    International Workshop on Semantic Evaluation (SemEval-2020) \\
    This work is licensed under a Creative Commons 
    Attribution 4.0 International Licence.
    Licence details:
    \url{http://creativecommons.org/licenses/by/4.0/}.
    %
    %
}

Offensive language is pervasive in social media in this day and age. Offensive language is so common and it is often used as emphasis instead of its semantic meaning, because of this, it can be hard to identify truly offensive content.  Individuals frequently take advantage of the perceived anonymity of computer-mediated communication, using this to engage in behaviour that many of them would not consider in real life. Online communities, social media platforms, and technology companies have been investing heavily in ways to cope with offensive language to prevent abusive behavior in social media. One of the most effective strategies for tackling this problem is to use computational methods to identify offense, aggression, and hate speech in user-generated content (e.g. posts, comments, microblogs, etc.).

The SemEval 2020 task on abuse detection \cite{2006.07235} aims to study both the target and the type of offensive language that has not been covered by previous works on various offenses such as hate speech detection and cyberbullying. In this paper, we focus on the first two sub-tasks: \emph{Sub-task A}, that focuses on offensive language or profanity detection, a binary classification problem wherein the objective is to determine if a tweet is offensive or not. \emph{Sub-task B}, which focuses on the identification of target presence, also a binary classification problem, wherein we are required to determine if a tweet is targeted at someone or something or not. 

\subsection{Boosting Pre-trained Representations}\label{section:intro-boosting}
Recent Natural Language Processing (NLP) systems have focused on the use of deep learning methods that take word embeddings as input. While these methods have been extremely successful on several tasks, we believe that information pertaining to the importance of individual words available at a corpus level might not be effectively captures by models that use pre-trained embeddings, especially given the small number of training epochs (usually 3) used.  We hypothesise that deep learning models, especially those that use pre-trained embeddings and so are trained on a small number of epochs, can benefit from corpus level count information. We test this on Sub-Task A using an ensemble of BERT and TF-IDF which outperforms both the individual models (Section \ref{results:subtaska}). 

For sub-task B, we hypothesise that these sentence representations can benefit from having POS information to help identify the presence of a target. To test this hypothesis, we integrate the count of part-of-speech (POS) tags with BERT. While this combination did outperform BERT, we found that a simpler modification to BERT (i.e. cost weighting, Section \ref{section:costweighting}) outperforms this combination.

\section{Related Work}


The Offensive Language Identification Dataset (OLID) was designed by \newcite{zampieri2019semeval} for the 2019 version of this task as there was  no prior work on this task before then. OLID is made up of 14,100 tweets that were annotated using experienced annotators but suffered from limited size, especially class imbalance. To get around this, OffensEval 2020 made use of the Semi-Supervised Offensive Language Identification Dataset (SOLID)~\cite{rosenthal2020large}.

\subsection{Prior OffensEval Systems}

Based on the results of OffensEval 2019, it seems that BERT is itself very powerful and it does relatively well for all of the 3 sub-tasks. In this section, we examine some of the best performing models on their techniques that we refer to for our methods.

\newcite{nikolov-radivchev-2019-nikolov} use a large variety of models and combined the best models in ensembles. They did pre-processing on the tweets by separating hashtag-ed tokens into separate words split by camel case. Stop words for the second and third sub-task were filtered because certain nouns and pronouns could contain useful information for the models to detect targets. 
Due to the class imbalance in the second and third sub-task, they used a variation of techniques to deal with this imbalance. They used oversampling by duplicating examples from the poorly represented classes. They also changed the class weights to provide more weight to the classes that are poorly represented. They also modified the thresholds that were used to classify an example, instead of having equal split for binary classes of 0.5, they shifted the boundary to accommodate the imbalance. Ensemble models were found to have over-fit the training data compared to BERT which had the best generalisation. Their BERT submissions were able to achieve $2^{nd}$ for the first sub-task and $1^{st}$ for the last sub-task. 

Similarly, work by \newcite{liu-etal-2019-nuli} mostly looked at pre-processing inputs before feeding it to BERT. It seems that pre-processing for BERT works very well in terms of improving its results. They also used hashtag segmentation, other techniques includes emoji substitution, they used a emoji Unicode to phrase Python library to increase semantic meaning in tweets. With just pre-processing alone they were able to achieve $1^{st}$ place for the first sub-task.

A significantly different method was used by \newcite{han-etal-2019-jhan014}, who used a rule based sentence offensiveness calculation to evaluate tweets. High and low offensive values are automatically classified as offensive or non-offensive, otherwise it follows a probabilistic distribution. For sub-task B using the sentence offensiveness model, they outperformed other systems that used deep learning or non-neural machine learning. This is a interesting find as it shows that traditional techniques such as using rule based models for target classification can be very successful compared to deep learning methods.

%
\section{System Overview}

For sub-task A, we test three models: a standard neural network that uses TF-IDF features, BERT and the ensemble of these two. For sub-task B we use noun counts, BERT and the ensemble of both.

\subsection{TF-IDF}

In order to incorporate global information into our model, we need to employ a technique that does so and TF-IDF does this well. Using TF-IDF, we will be able to identify keywords that helps us to distinguish between offensive/non-offensive tweets which offensive tweets will tend to have more offensive words while non-offensive tweets usually contains more neutral-toned words. Since we use TF-IDF as our input features to be combined with BERT, we have a neural network so that when we are training the combination of the models, the neural network will enable us to maintain learning from training compared to non-neural machine learning techniques.

\subsection{BERT}

The reason for picking BERT \cite{DBLP:journals/corr/abs-1810-04805} is because BERT has outperformed several similar techniques that provides sentence level embeddings such as BiLSTM and ELMo \cite{Peters:2018}. It has also shown to be very effective at doing all the sub-tasks in the previous year evaluation \cite{zampieri2019semeval}. We can see that it has both strengths in generalisation and also able to handle contextual based evaluations well. 

\subsection{Ensemble Model}

Ensemble techniques have shown to be effective in reducing variance in the prediction and at making better predictions, this can be achieved for neural networks having multiple sources of information \cite{brownlee2018better}. We will be using an ensemble model to combine individual models into one. Just using BERT alone will provide us sentence level information, but if we combine BERT features and TF-IDF features, we can have access to both sentence and corpus level information which is the goal of our hypothesis. This ensemble model is created by concatenating the sentence representation of BERT to the features generated by the TF-IDF model before then using this combined vector for classification. In practice, this translates into calculating the TF-IDF vector for each sentence and concatenating it to the corresponding BERT output. This vector is then fed to a fully connected classification layer. Both BERT and the TF-IDF weights are updated during training. 

\subsection{Noun Count As Features}

We have seen the success of rule based method for sub-task B that achieved significant performance compared to machine learning techniques. \cite{han-etal-2019-jhan014} has shown that using a manually annotated offensive list of words that provides a measure of the strength of offensiveness is effective. Since targets are very likely to be identified as nouns or pronouns in the tweets, we can identify the presence of a target if we have a count of part-of-speech tags such as `PRP', `NP'. 

\subsection{Cost Weight Adjustments}
\label{section:costweighting}
An analysis of the datasets showed that for all sub-tasks, there were large class imbalances. We follow the method described by \newcite{tayyar-madabushi-etal-2019-cost} to modify the cost function to allow poorly represented classes to have more impact when calculating the cost of error. They show that other techniques such as data augmentation through oversampling does not improve the performance of BERT. We use cost weighting for both tasks. 

\begin{table}[ht]
\centering
\begin{tabular}{|l|l|l|l|l|} 
\hline
Sub-task & Total                                        & Class 1                                             & Class 2                                        \\ 
\hline
A        & \textcolor[rgb]{0.129,0.129,0.129}{9075418 } & \textcolor[rgb]{0.129,0.129,0.129}{1446768 (15\%)~} & \textcolor[rgb]{0.129,0.129,0.129}{7628650 (85\%)}                                                \\ 
\hline
B        & \textcolor[rgb]{0.129,0.129,0.129}{188974 }  & \textcolor[rgb]{0.129,0.129,0.129}{39424 (20\%)}    & \textcolor[rgb]{0.129,0.129,0.129}{149550 (80\%)}                                            \\ 
\hline
\end{tabular}
\caption{Class Imbalance Analysis}
\end{table}

\section{Experimental Setup}

For each of the sub-tasks we participate in, we split our training set into a training set and development set in a 4:1 ratio. Our test set is the evaluation set for SemEval-2019. We submit the best version of these experiments to SemEval-2020. Also, in each case, we first experiment with BERT, then by adding additional parameters to BERT and finally by use of cost-weights. All ensemble models were created at the embedding layer by appending additional features to BERT embeddings before then using a fully connected layer for classification. 

\subsection{Sub-Task A}

Our setup for sub-task A was to pre-process using stemming and NLTK's tweet tokenizer for the TF-IDF features, where we only consider the top 6000 highest term frequency words to accommodate memory limitations. With BERT, we found that stemming or lemmatization does not help to improve the results, so our input uses BERT's default tokenizer with a maximum sequence length of 64. We used the English dataset provided for training, which consists of nine million examples. Unfortunately, due to memory constraints, we were unable to use this entire dataset for training and the final model used just 10\% of this dataset. We also applied cost weighting to account for class imbalances. We used a learning rate of 5e-6, batch size of 32. We found the best results to be within 1 to 2 epochs.

\subsection{Sub-Task B}

Our setup for sub-task B was using the OLID dataset, as we found that the new dataset provided had a high rate of misclassified label. We used NLTK's POS tagger to extract the tags for our noun count features and we extract the count of `NNS' and `PRP' tags as they give the most information about target presence. We used a learning rate of 5e-5, batch size of 32. We found the best results to be within 20 epochs as the dataset is small, we needed to adjust for the step size decrease.

\section{Results and Analysis}
We present our overall rankings on each of the two sub-tasks in Table \ref{table:results}. While our rank on the first task is not very high, we note that it is within 2 points of the top scoring team - it should be emphasised that we achieve this result by use of only 10\% of the the available training data due to GPU memory limitations. We rank much close to the top on Sub-Task 2 with a rank of 4 amongst a total of 44 submissions. 
\begin{table}[h]
    \begin{subtable}[t]{.48\linewidth}
        \centering
            \begin{tabular}{|@{\extracolsep{1pt}}lll|}
                \hline
                Rank & System & Macro F1 \\ 
                \hline
                1 & ltuhh2020 & 0.92226 \\ 
                2 & gwiedemann & 0.92040 \\ 
                3 & Galileo & 0.91985 \\ 
                \multicolumn{3}{|c|}{\dots} \\
                38 & \textbf{wml754 (This work)} & \textbf{0.90901} \\ 
                \hdashline
                & All NOT & 0.4193 \\ 
                & All OFF & 0.2174 \\
                \hline 
            \end{tabular}
    \caption{Sub-task A}
    \end{subtable}
    \begin{subtable}[t]{.48\linewidth}
        \centering
        \begin{tabular}{|@{\extracolsep{1pt}}lll|}
            \hline
                Rank & System & Macro F1 \\ 
                \hline
                1 & Galileo & 0.74618 \\ 
                2 & tracypg & 0.73623 \\ 
                3 & pochunchen & 0.69063 \\ 
                4 & \textbf{wml754 (This Work)} & \textbf{0.67336} \\ 
                \hdashline
                & All TIN & 0.3741 \\ 
                & All UNT & 0.2869 \\
                \hline
        \end{tabular}
    \caption{Sub-task B}
    \end{subtable}
    
    \caption{\label{table:results}Rankings on Sub-Task 1 and Sub-Task 2}
    
\end{table}

\subsection{Sub-Task A}
As described in Section \ref{section:intro-boosting}, we hypothesise that deep learning models, especially those that use pre-trained embeddings and so are trained on a small number of epochs, can benefit from corpus level count information. So as to test this hypothesis we create three different models: A Simple Neural Network (SNN), consisting of a single layer, that is fed with TF-IDF features, BERT and an ensemble of the two. The ensemble model is constructed by removing the classification layer from BERT and the SNN model, concatenating their output and passing this concatenated vector through a new fully connected layer for classification. We perform our experiments using 10\% of the training data as our training set, and last year's SemEval test set as our test set. We pick the best performing model and use that to generate results for our submission. Unfortunately we were unable to train on more than 10\% of the training set. Our experiments show that corpus level count information captured by TF-IDF can indeed boost the performance of BERT. Table \ref{table:experiments-a-noweights} details the results of our experiments with the three models.
\label{results:subtaska}
\begin{table}[ht]
\centering
\begin{tabular}{|l|r|r|r|r|r|r|} 
\hline
Model & \multicolumn{1}{l|}{Macro F1 (Train)} & \multicolumn{1}{l|}{Macro F1 (Dev)} \\ 
\hline
SNN & 0.9227 & 0.7329 \\ 
BERT & 0.9641 & 0.7378 \\ 
Ensemble & 0.9179 & \textbf{0.7819} \\
\hline
\end{tabular}
\caption{A comparison of a Simple Neural Network (SNN) fed with TF-IDF features, BERT and the ensemble of the two. }\label{table:experiments-a-noweights}
\end{table}

Our analysis of the training and test data using the Wilcoxon signed-rank test as described by \newcite{tayyar-madabushi-etal-2019-cost} shows that the training and development sets are different enough to warrant the use of cost-weighting (Section \ref{section:costweighting}). To this end we introduce cost weighting to each of the three models described above and the results of these experiments are presented in Table \ref{table:experiments-a-yesweights}.

\begin{table}[ht]
\centering
\arrayrulecolor{black}
\begin{tabular}{|l|r|r|r|r|r|r|r|r|} 
\arrayrulecolor[rgb]{0.8,0.8,0.8}\cline{1-1}\arrayrulecolor{black}\cline{2-9}
\multicolumn{1}{!{\color[rgb]{0.8,0.8,0.8}\vrule}l|}{} & \multicolumn{2}{l|}{Cost Weight} & \multicolumn{3}{l|}{Train} & \multicolumn{3}{l|}{Dev} \\ 
\hline
Model & \multicolumn{1}{l|}{OFF} & \multicolumn{1}{l|}{NOT} & \multicolumn{1}{l|}{Precision} & \multicolumn{1}{l|}{Recall} & \multicolumn{1}{l|}{F1} & \multicolumn{1}{l|}{Precision} & \multicolumn{1}{l|}{Recall} & \multicolumn{1}{l|}{F1} \\ 
\hline
SNN & 10 & 1 & 0.8233 & 0.8923 & 0.8512 & 0.7619 & 0.7448 & 0.7523 \\ 
\hline
BERT & 50 & 1 & 0.9267 & 0.9615 & 0.9430 & 0.8123 & 0.8050 & 0.8085 \\ 
\hline
 Ensemble & 100 & 1 & 0.8896 & 0.9604 & 0.9197 & 0.8095 & 0.8165 & \textbf{0.8128} \\
\hline
\end{tabular}
\caption{A Cost Weight Adjusted comparison of a Simple Neural Network (SNN) fed with TF-IDF features, BERT and the ensemble of the two.}\label{table:experiments-a-yesweights}
\end{table}

We observe that adding the optimal cost weights to poorly represented classes significantly improves the performance of all models. The ensemble model, however, still outperforms either of SNN or BERT, despite a large increase in performance for BERT after adding cost weights.

As mentioned we were only able to train our BERT and ensemble models with 10\% of the training data. The performance of our models can be further improved given GPU resources as shown in Table \ref{table:needgpu}.

\begin{table}[ht]
\centering
\arrayrulecolor{black}
\begin{tabular}{|r|r|r|r|r|r|r|} 
\arrayrulecolor[rgb]{0.8,0.8,0.8}\cline{1-1}\arrayrulecolor{black}\cline{2-7}
\multicolumn{1}{!{\color[rgb]{0.8,0.8,0.8}\vrule}l|}{} & \multicolumn{3}{l|}{Train} & \multicolumn{3}{l|}{Dev} \\ 
\hline
\multicolumn{1}{|l|}{Data Size} & \multicolumn{1}{l|}{Precision} & \multicolumn{1}{l|}{Recall} & \multicolumn{1}{l|}{F1} & \multicolumn{1}{l|}{Precision} & \multicolumn{1}{l|}{Recall} & \multicolumn{1}{l|}{F1} \\ 
\hline
100000 & 0.8896 & 0.9604 & 0.9197 & 0.8095 & 0.8165 & 0.8128 \\ 
\hline
800000 & 0.9652 & 0.9413 & 0.9527 & 0.8364 & 0.8095 & \textbf{0.8212} \\
\hline
\end{tabular}
\caption{We show that our models can perform better if trained on more of the training data.} \label{table:needgpu}
\end{table}

\subsection{Sub-Task B}
\label{results:subtaskb}

For sub-task B, as mentioned in Section \ref{section:intro-boosting}, we hypothesise that these sentence representations can benefit from having POS information to help identify the presence of a target. To test this hypothesis, we integrate the count of part-of-speech (POS) tags with BERT. We use the OLID dataset for training and last year's evaluation set as the test set. The best performing model is used to make predictions for submission to this year's competition. We present the results of our experiments in Table \ref{table:experiments-b}. Our experiments show that while noun counts do improve the accuracy of BERT, cost weighting BERT is more effective. 

\begin{table}[h!]
\centering
\begin{tabular}{|l|l|l|l|l|l|l|l|l|} 
\cline{2-9}
\multicolumn{1}{l|}{} & \multicolumn{2}{l|}{Cost Weight} & \multicolumn{3}{l|}{Train} & \multicolumn{3}{l|}{Dev} \\ 
\hline
Model & TIN & UNT & Precision & Recall & F1 & Precision & Recall & F1 \\ 
\hline
BERT & 1 & 1 & 0.8285 & 0.7582 & 0.7875 & 0.7533 & 0.6849 & 0.7115 \\ 
\hline
BERT & 1 & 4 & 0.8283 & 0.7347 & 0.7707 & 0.7604 & 0.7520 & \textbf{0.7561} \\ 
\hline
BERT + NC & 1 & 1 & 0.9177 & 0.8805 & 0.8979 & 0.7504 & 0.7173 & 0.7321 \\ 
\hline
BERT + NC & 1 & 4 & 0.8043 & 0.7767 & 0.7896 & 0.7175 & 0.8026 & 0.7485 \\
\hline
\end{tabular}
\caption{Comparison before and after adding Noun Count}
\label{table:experiments-b}
\end{table}

\section{Conclusion}

We show that incorporating corpus level information does help improve the performance of BERT. We achieve competitive results using just 10\% of the available dataset and would like to test the limits by training with the full dataset. Our experiments also show that noun counts do help boost the performance of BERT, but not as much as cost-weighting. 


\bibliographystyle{coling}
\bibliography{semeval2020}

\begin{thebibliography}{}

\bibitem[\protect\citename{Brownlee}2018]{brownlee2018better}
J.~Brownlee.
\newblock 2018.
\newblock {\em Better Deep Learning: Train Faster, Reduce Overfitting, and Make
  Better Predictions}.
\newblock Machine Learning Mastery.

\bibitem[\protect\citename{Devlin \bgroup et al.\egroup
  }2018]{DBLP:journals/corr/abs-1810-04805}
Jacob Devlin, Ming{-}Wei Chang, Kenton Lee, and Kristina Toutanova.
\newblock 2018.
\newblock {BERT:} pre-training of deep bidirectional transformers for language
  understanding.
\newblock {\em CoRR}, abs/1810.04805.

\bibitem[\protect\citename{Han \bgroup et al.\egroup
  }2019]{han-etal-2019-jhan014}
Jiahui Han, Shengtan Wu, and Xinyu Liu.
\newblock 2019.
\newblock jhan014 at {S}em{E}val-2019 task 6: Identifying and categorizing
  offensive language in social media.
\newblock In {\em Proceedings of the 13th International Workshop on Semantic
  Evaluation}, pages 652--656, Minneapolis, Minnesota, USA, June. Association
  for Computational Linguistics.

\bibitem[\protect\citename{Liu \bgroup et al.\egroup }2019]{liu-etal-2019-nuli}
Ping Liu, Wen Li, and Liang Zou.
\newblock 2019.
\newblock {NULI} at {S}em{E}val-2019 task 6: Transfer learning for offensive
  language detection using bidirectional transformers.
\newblock In {\em Proceedings of the 13th International Workshop on Semantic
  Evaluation}, pages 87--91, Minneapolis, Minnesota, USA, June. Association for
  Computational Linguistics.

\bibitem[\protect\citename{Nikolov and
  Radivchev}2019]{nikolov-radivchev-2019-nikolov}
Alex Nikolov and Victor Radivchev.
\newblock 2019.
\newblock Nikolov-radivchev at {S}em{E}val-2019 task 6: Offensive tweet
  classification with {BERT} and ensembles.
\newblock In {\em Proceedings of the 13th International Workshop on Semantic
  Evaluation}, pages 691--695, Minneapolis, Minnesota, USA, June. Association
  for Computational Linguistics.

\bibitem[\protect\citename{Peters \bgroup et al.\egroup }2018]{Peters:2018}
Matthew~E. Peters, Mark Neumann, Mohit Iyyer, Matt Gardner, Christopher Clark,
  Kenton Lee, and Luke Zettlemoyer.
\newblock 2018.
\newblock Deep contextualized word representations.
\newblock In {\em Proc. of NAACL}.

\bibitem[\protect\citename{Rosenthal \bgroup et al.\egroup
  }2020]{rosenthal2020large}
Sara Rosenthal, Pepa Atanasova, Georgi Karadzhov, Marcos Zampieri, and Preslav
  Nakov.
\newblock 2020.
\newblock A large-scale semi-supervised dataset for offensive language
  identification.
\newblock {\em arXiv preprint arXiv:2004.14454}.

\bibitem[\protect\citename{Tayyar~Madabushi \bgroup et al.\egroup
  }2019]{tayyar-madabushi-etal-2019-cost}
Harish Tayyar~Madabushi, Elena Kochkina, and Michael Castelle.
\newblock 2019.
\newblock Cost-sensitive {BERT} for generalisable sentence classification on
  imbalanced data.
\newblock In {\em Proceedings of the Second Workshop on Natural Language
  Processing for Internet Freedom: Censorship, Disinformation, and Propaganda},
  pages 125--134, Hong Kong, China, November. Association for Computational
  Linguistics.

\bibitem[\protect\citename{Zampieri \bgroup et al.\egroup
  }2019]{zampieri2019semeval}
Marcos Zampieri, Shervin Malmasi, Preslav Nakov, Sara Rosenthal, Noura Farra,
  and Ritesh Kumar.
\newblock 2019.
\newblock Semeval-2019 task 6: Identifying and categorizing offensive language
  in social media (offenseval).
\newblock {\em arXiv preprint arXiv:1903.08983}.

\bibitem[\protect\citename{Zampieri \bgroup et al.\egroup }2020]{2006.07235}
Marcos Zampieri, Preslav Nakov, Sara Rosenthal, Pepa Atanasova, Georgi
  Karadzhov, Hamdy Mubarak, Leon Derczynski, Zeses Pitenis, and Çağrı
  Çöltekin.
\newblock 2020.
\newblock Semeval-2020 task 12: Multilingual offensive language identification
  in social media (offenseval 2020).

\end{thebibliography}

\end{document}